\documentclass{egpubl}
\usepackage{pg2026}

\SpecialIssuePaper         

\CGFStandardLicense

\usepackage[T1]{fontenc}
\usepackage{dfadobe}  
\usepackage{amssymb}
\usepackage{booktabs}
\usepackage{amsmath}
\usepackage[table,xcdraw]{xcolor}
\usepackage{multirow}
\usepackage{cite}  
\BibtexOrBiblatex
\electronicVersion
\PrintedOrElectronic
\usepackage{graphicx}
\usepackage{egweblnk} 

\title[LiteTex-GS]%
      { LiteTex-GS: Fast and Lightweight Texturing for Gaussian Splatting}

\author[Li and Guo et al.]
{
{\parbox{\textwidth}{\centering
Zhiwei Li\(^{1,}\)
\thanks{%
\raggedright
These authors contributed equally.\\[-1pt]
\(^\ast\) Corresponding authors:
\texttt{raohong@ncu.edu.cn}; 
\texttt{chenshengbo@ncu.edu.cn};
\texttt{lei.ma@pku.edu.cn}.
}
\orcid{0009-0006-9422-185X},
Yijia Guo\(^{2,\dagger}\)\orcid{0000-0001-8588-4498},
Yishi Lu\(^{3}\)\orcid{0009-0005-1374-4759},
Liwen Hu\(^{2}\)\orcid{0000-0002-0257-5292},
Hong Rao\(^{4,\ast}\)\orcid{0000-0002-7467-7108},
Shengbo Chen\(^{5,\ast}\)\orcid{0000-0002-9026-8128},
Lei Ma\(^{2,6,7,\ast}\)\orcid{0000-0001-6024-3854}
}}\\
{\parbox{\textwidth}{\centering
\(^1\)School of Mathematics and Computer Science, Nanchang University\\
\(^2\)State Key Laboratory of Multimedia Information Processing,
School of Computer Science, Peking University\\
\(^3\)Henan University 
\(^4\)School of Software, Nanchang University
\(^5\)School of Artificial Intelligence, Nanchang University\\
\(^6\)National Biomedical Imaging Center, Peking University
\(^7\)College of Future Technology, Peking University
}}
}

\begin{document}

\teaser{
 \vspace{-3mm}
 \includegraphics[width=0.9\linewidth]{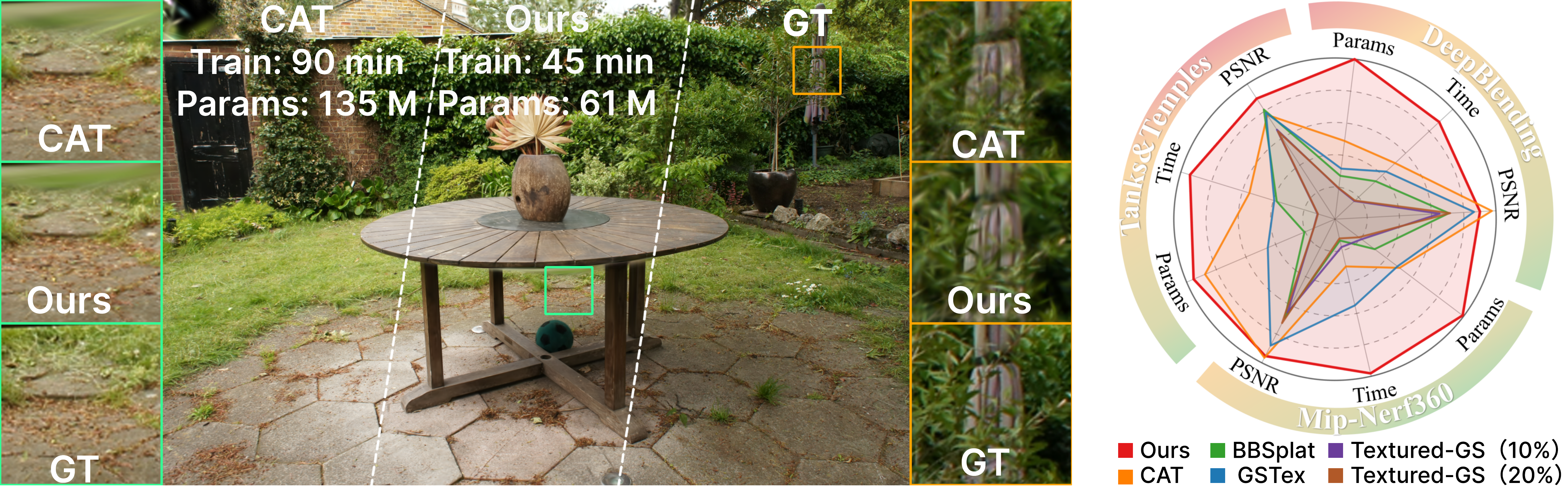}
 \centering
\caption{
Left: Qualitative comparison on a representative Mip-NeRF 360 scene. Two zoom-in regions compare LiteTex-GS mainly with CAT, with ground truth as reference; the average training time and total parameter count on Mip-NeRF 360 are annotated below each method. Right: Quality--efficiency trade-off in terms of PSNR, training time, and parameter count. LiteTex-GS maintains competitive visual fidelity while substantially reducing both parameters and training time through global-to-local capacity allocation.
}
\label{fig:teaser}
}

\maketitle

\begin{abstract}
Gaussian Splatting has enabled real-time novel view synthesis, but its tightly coupled geometry and appearance representation often require a large number of primitives to reproduce high-frequency texture details, leading to substantial memory and optimization costs. Recent textured 2D Gaussian methods alleviate this limitation by attaching texture maps to Gaussian primitives. However, bridging the fundamental structural gap between discrete Gaussians and continuous 2D grids requires complex parameterizations that introduce severe computational overhead. This overhead fundamentally compromises the original efficiency of Gaussian Splatting, making the balance between detailed texturing and computational agility an unresolved challenge.
To address these challenges, we propose LiteTex-GS, a fast and lightweight texturing framework for Gaussian Splatting. Our method initializes an extremely compact representation—assigning minimal local texture to each Gaussian and progressively allocates higher resolution only to primitives with significant reconstruction errors. To maintain a streamlined geometric scaffold, we introduce a contribution- and area-aware pruning strategy that eliminates low-utility Gaussians. Furthermore, to mitigate the gradient dilution caused by texture upsampling, we design a resolution-aware update rule that preserves rapid and stable convergence. Extensive experiments on standard novel view synthesis benchmarks demonstrate that our method achieves competitive or superior rendering quality while using substantially fewer parameters and less training time than existing textured Gaussian baselines.

\ccsdesc[500]{Computing methodologies~Computer graphics~Rendering}
\ccsdesc[300]{Computing methodologies~Artificial intelligence~Computer vision}

\printccsdesc

\end{abstract}  


\section{Introduction}
\label{sec:intro}

While Gaussian Splatting \cite{kerbl20233d,huang20242d} has emerged as a powerful paradigm for real-time, high-fidelity novel view synthesis, its explicit, point-based formulation suffers from severe representation inefficiencies when modeling complex scenes. To faithfully capture rich, high-frequency textures, standard 3DGS inherently necessitates an exceedingly massive number of primitives, resulting in prohibitive memory and storage bottlenecks. This unconstrained proliferation is fundamentally rooted in the tightly entangled representation of geometry and appearance within the standard framework. Lacking a principled structural disentanglement between the underlying spatial scaffold and surface texture, the optimization process is forced to aggressively densify and instantiate redundant geometric primitives purely to express intricate color transitions. 

To explicitly decouple geometry from appearance, recent literature has explored various mechanisms to augment the splatting pipeline with auxiliary textural representations. Most of these methods\cite{chao2025textured,rong2025gstex,svitov2025billboard,papantonakis2025content} use planar 2D Gaussians to match the dimensionality of traditional textures. To bridge the fundamental structural gap between these unstructured, discrete spatial elements and continuous 2D coordinate grids, such frameworks must establish a dense mapping, typically through complex UV parameterization or continuous coordinate transformations. However, enforcing consistent and seamless parameterization across millions of dynamically optimized, disjointed primitives is inherently ill-posed and prone to a severe computational overhead, ultimately compromising the reconstruction efficiency that is the hallmark of the original splatting architecture.  Consequently, striking a principled balance between expressive, disentangled high-frequency detail and the intrinsic computational agility of the splatting paradigm remains a critical challenge.

To overcome these limitations, we propose LiteTex-GS, a texture-decoupled Gaussian splatting framework that balances expressive detail with computational efficiency through unified global-to-local capacity allocation. We assign independent local textures to Gaussians and initialize them with compact low-resolution maps. Globally, a frequency-aware scheduler jointly controls image supervision resolution, the texture-resolution cap, and the primitive densification rate, guiding training from coarse supervision to finer texture allocation. Locally, unlike prior textured Gaussian methods~\cite{papantonakis2025content,chao2025textured} that statically allocate texture resolution from geometric priors, LiteTex-GS increases only the personal resolution ceilings of high-error primitives under the current global cap, so newly available texels are consumed selectively in visually difficult regions. To keep the geometric scaffold compact, we further use contribution- and area-aware geometric management, motivated by the limitation of opacity-based pruning in standard 3DGS~\cite{kerbl20233d}. Finally, to mitigate gradient dilution after texture upsampling, we introduce a resolution-aware update rule for the texture branch. While existing methods~\cite{chao2025textured,papantonakis2025content,xie2025fact} use update scaling for static pre-allocation, our rule adapts the effective update magnitude as texture resolutions evolve, improving the optimization of newly allocated high-resolution textures within the original training budget. We summarize our main contributions as follows:
\begin{itemize}
    \item A global-to-local capacity allocation framework for lightweight textured Gaussian splatting. A DashGaussian-based frequency-aware scheduler controls rendering resolution, texture-cap release, and primitive densification, while per-Gaussian reconstruction errors determine where the newly available texture capacity is locally consumed.
    \item An adaptive texture-growth and geometry-management strategy that selectively increases texture resolution for high-error primitives and removes persistently low-utility geometry, producing a compact representation with important local structures and texture details.
    \item A resolution-aware texture optimization rule that compensates for the optimization imbalance caused by progressive texture growth, improving convergence and the quality--efficiency trade-off.
\end{itemize}

\noindent Our approach achieves competitive rendering quality while using substantially fewer parameters and shorter training time than existing textured Gaussian baselines, demonstrating an efficient quality--efficiency trade-off.

\section{Related Work}
\label{gen_inst}

\subsection{Representations for 3D Reconstruction and Textured Gaussian Splatting}

Early 3D reconstruction methods, such as Multi-Plane Images (MPIs) \cite{zhou2018stereo,shade1998layered}, adopt layered textured planes for efficient rendering but lack sufficient geometric expressiveness. Neural Radiance Fields (NeRF) \cite{mildenhall2021nerf,guo2024spike,xiang2021neutex,fridovich2022plenoxels,chen2022tensorf,muller2022instant} popularized implicit volumetric representations, with subsequent works enhancing their speed, scalability, and detail capture capability; however, they still lag behind Gaussian Splatting in real-time performance. 3D Gaussian Splatting \cite{kerbl20233d,hahlbohm2026faster,hanson2025speedy,fan2024lightgaussian} has become the dominant approach for real-time novel view synthesis due to its efficient rasterization pipeline, while 2D Gaussian Splatting \cite{huang20242d,zhang2024gaussianimage,yang2025introducing} further improves geometric accuracy by flattening primitives into planar surfels. Nevertheless, standard Gaussian Splatting binds a single color to each primitive, necessitating excessive tiny primitives to represent fine textures and leading to redundant parameters and high storage costs. To address the texture representation limitation of standard Gaussian Splatting, recent works have introduced textured Gaussian primitives to decouple appearance and geometry. BBSplat and Textured Gaussians assign fixed-resolution texture to each Gaussian primitive \cite{svitov2025billboard, chao2025textured}, enabling finer appearance detail capture but lacking adaptive resource allocation, which results in either insufficient detail or redundant parameter storage. GSTex \cite{rong2025gstex} improves upon this by distributing texels based on the size of each primitive, but it relies on heuristic-based texel allocation and suffers from texture distortion during primitive scaling. Content-Aware Texturing (CAT) \cite{papantonakis2025content} further improves texture allocation by considering local scene frequency and world-space texel size, which helps reduce scaling artifacts and improves the distribution of texture capacity. However, its allocation strategy is still primarily tied to spatial and geometric heuristics. 

\subsection{Efficient Optimization and Scheduling}

Optimization efficiency is crucial for the practical deployment of Gaussian Splatting-based methods. Early acceleration methods\cite{kerbl20233d,hahlbohm2026faster,hanson2025speedy,wang2025faster,guo2024prtgs,lu2024scaffold} focus on engineering optimizations, such as refining forward/backward pipelines and employing sparse optimizers to reduce per-iteration costs; other works \cite{kerbl20233d,hanson2025pup,lee2026safeguardgs} accelerate training by pruning redundant primitives, often at the cost of degraded rendering quality. Primitive densification control has also been explored: Taming-3DGS\cite{mallick2024taming} and Revising-3DGS\cite{rota2024revising} regulate primitive growth with predefined budgets but depend heavily on manual hyperparameter settings, while level-of-detail (LoD) methods\cite{kerbl2024hierarchical,cheng2025clod,kulhanek2025lodge,guo2025fly} use multi-resolution rendering to balance efficiency and quality but introduce additional training complexity. DashGaussian~\cite{chen2025dashgaussian} introduces a frequency-aware
coarse-to-fine scheduling strategy that progressively increases rendering
resolution and coordinates primitive growth according to the frequency
content of the training images. We adopt this scheduling principle as the
global curriculum for our textured Gaussian framework, where it is further
used to regulate texture-cap release. In summary, existing Gaussian Splatting-based methods face prominent bottlenecks in balancing optimization efficiency and rendering quality. Optimization scheduling strategies either sacrifice quality, rely on manual tuning, introduce extra complexity, or fail to adapt to textured representations. Our work addresses these challenges from a complementary perspective by jointly coordinating rendering resolution, adaptive texture allocation, and primitive pruning for efficient textured Gaussian splatting.

\section{Method}

\begin{figure*}[t]
  \centering
  \includegraphics[width=1\linewidth]{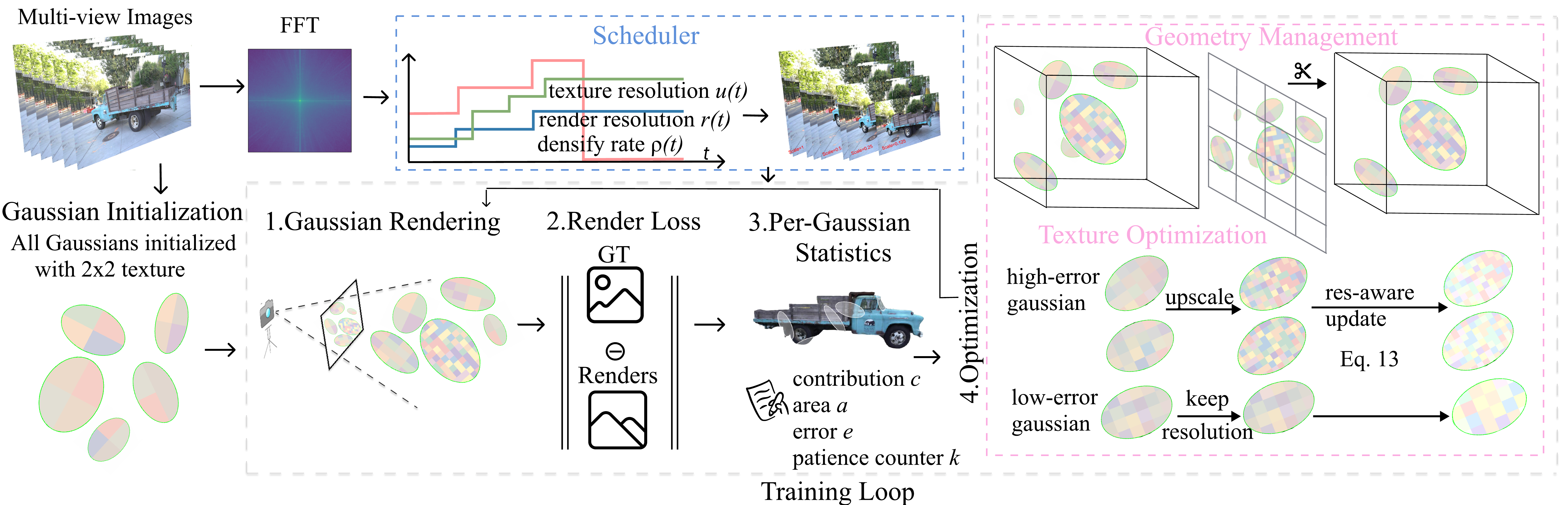}
  \caption{Overview of our proposed framework. Guided by a global frequency-aware scheduler, our pipeline optimizes scenes in a principled coarse-to-fine manner. Starting from ultra-compact $2\times2$ local textures, we tightly couple contribution- and area-aware geometric pruning with error-driven texture upsampling. A resolution-aware update scaling rule is further applied to mitigate gradient dilution and stabilize the optimization of upsampled textures.}
  \label{pipeline}
\end{figure*}

Our method extends textured Gaussian splatting with a content-aware training strategy that jointly controls rendering resolution, per-primitive texture resolution, and primitive management. An overview of our framework is illustrated in Figure~\ref{pipeline}. We represent a scene as a compact set of 2D Gaussian primitives, where geometry and appearance are decoupled. Geometry is parameterized by the standard Gaussian attributes, while appearance is stored in a learnable local texture map attached to each primitive. Unlike fixed appearance embeddings, these texture maps can change their resolution during optimization. This allows the model to keep easy regions compact while allocating substantially more appearance capacity to visually difficult regions. 

\subsection{Per-Primitive Texture Representation}

Each Gaussian primitive is parameterized by its position, rotation, scale, and opacity, together with a local RGB texture map $\mathcal{T}_i$. The texture is defined in the local coordinate frame of the primitive, which makes the representation scale-independent: when the primitive changes size during optimization, the texture remains attached to the same local support instead of being stretched in the global image domain. For a pixel $\mathbf{p}$, the rendered color is obtained by compositing the contributions of the visible primitives and sampling their local textures:
\begin{equation}
\hat{I}(\mathbf{p}) =
\sum_{i \in \mathcal{N}(\mathbf{p})}
\omega_i(\mathbf{p})\,
\mathcal{S}\!\left(\mathcal{T}_i, \Phi_i(\mathbf{p})\right),
\end{equation}
where $\mathcal{N}(\mathbf{p})$ denotes the set of primitives that influence pixel $\mathbf{p}$, $\omega_i(\mathbf{p})$ is the usual visibility-and-opacity-based blending weight, $\Phi_i(\mathbf{p})$ maps the pixel to the local texture coordinates of primitive $i$, and $\mathcal{S}$ is differentiable bilinear sampling~\cite{jaderberg2015spatial}.

More specifically, $\Phi_i(\mathbf{p})$ is obtained by intersecting the camera ray of pixel $\mathbf{p}$ with the local tangent plane of Gaussian $i$, transforming the intersection point from view space to world space, and then projecting it into the primitive's axis-aligned local frame. The resulting canonical coordinate is normalized by the primitive's two-dimensional scale and converted to texture coordinates according to the current texel size and texture center offset. This mapping attaches the texture to the primitive support itself, rather than to the screen or to a global UV atlas. Pixels whose mapped texture coordinates fall outside the valid texture domain are marked as out of bounds and do not use texture sampling; we do not apply wrapping or padding. The opacity of each primitive is stored separately from the texture, so the local texture contains only RGB appearance values.

We initialize the texture branch in a deliberately compact state. Each primitive starts from a very small $2 \times 2$ texture map. During training, the active texture resolution is dynamically adjusted according to the primitive's canonical support, texel size, and personal resolution cap. The resolution is not restricted to a fixed global size and can be rectangular, which allows different primitives to allocate texture capacity according to their local geometric extent and reconstruction difficulty. This keeps the early optimization stable and lightweight, while leaving room for the model to allocate more capacity later once it has enough evidence from the rendered training views.

\subsection{Frequency-Aware Scheduling}
Building on the frequency-aware scheduling principle introduced by
DashGaussian~\cite{chen2025dashgaussian}, we extend the scheduler to coordinate image supervision, texture-cap release, and primitive densification during training. Rather than using a fixed iteration-based curriculum~\cite{cheng2025clod,tang2025hisplat,karras2017progressive}, we derive a scene-adaptive coarse-to-fine schedule from the frequency content of the training images. The intuition is that scenes with richer high-frequency content should transition more gradually to full-resolution supervision, whereas smoother scenes can advance more aggressively.

We first construct a frequency-significance image once before training. For each training image $I_m$, we compute a Sobel-based weight map $W_m$ to emphasize high-texture regions and foreground-like structures:
\begin{equation}
G_m =
\sqrt{
\left(Sobel_x * \bar I_m\right)^2
+
\left(Sobel_y * \bar I_m\right)^2
},
\end{equation}
\begin{equation}
W_m =
\mathrm{clip}
\left(
w_{\min} + \alpha \widetilde{G}_m,
w_{\min},
w_{\max}
\right),
\end{equation}
where $\bar I_m$ is the grayscale image, $\tilde G_m$ is the min--max normalized gradient magnitude, and we use $\alpha=1.5$, $w_{\min}=1.0$, and $w_{\max}=3.0$ by default. The weighted images are transformed to the frequency domain and averaged over all training views:
\begin{equation}
F_{\mathrm{img}} =
\frac{1}{M}
\sum_{m=1}^{M}
\left|
\mathrm{fftshift}
\left(
\mathcal{F}(W_m \odot I_m)
\right)
\right|.
\end{equation}
where $\mathcal{F}$ denotes the two-dimensional Fourier transform. This preprocessing is performed only once and introduces negligible overhead compared with iterative optimization.

From $\mathcal{F}_{\mathrm{img}}$, we generate a set of scene-specific frequency stages. Let $S(\mathcal{F}_{\mathrm{img}},s)$ denote the accumulated energy within the centered low-frequency window of size $(H/s)\times(W/s)$, where $s$ is a render downsampling factor. We determine the coarsest scale from a start significance factor $\beta$ and a configured maximum downsampling factor $s_{\mathrm{cfg}}$:
\begin{equation}
E_{\min} =
\frac{E_{\mathrm{total}}}{\beta},
\qquad
E_{\mathrm{total}} =
\sum_{u,v}
F_{\mathrm{img}}(u,v).
\end{equation}
In our experiments, we set $\beta=3$ and $s_{\mathrm{cfg}}=4$, so the coarsest image supervision is at most $1/4$ of the original resolution in each spatial dimension. We then sample at most $L=32$ frequency-significance targets between $E_{\min}$ and $E_{\mathrm{total}}$, map each target to a render scale using a binary-search scale solver, and merge adjacent stages that produce the same integer scale. The resulting number of stages is therefore scene-dependent.

At iteration $t$, the scheduler returns
\begin{equation}
(r(t), u(t), \rho(t))
=
\mathcal{S}(t; F_{\mathrm{img}}),
\end{equation}
where $r(t)$ is the render downsampling factor, $u(t)$ is the current cap on per-Gaussian texture resolution, and $\rho(t)$ is the densification rate. These three quantities are coordinated by the same frequency-aware stage schedule but control different aspects of training. Specifically, $r(t)$ determines the resolution of image supervision, $u(t)$ constrains the maximum texture resolution that a primitive can reach through error-driven growth, and $\rho(t)$ controls the rate of primitive densification. Thus, image resolution, texture capacity, and geometric growth follow a shared coarse-to-fine progression without being tied to the same numerical value.

During early training, the scheduler uses low-resolution supervision, a tight texture cap, and conservative densification to stabilize the geometric scaffold. As training advances, finer image supervision is activated, the texture resolution constraint is gradually relaxed, and densification is allowed to introduce more primitives. The detailed stage construction and scale interpolation are provided in the supplementary material.

\begin{figure}[htbp]
    \centering
    \includegraphics[width=\linewidth]{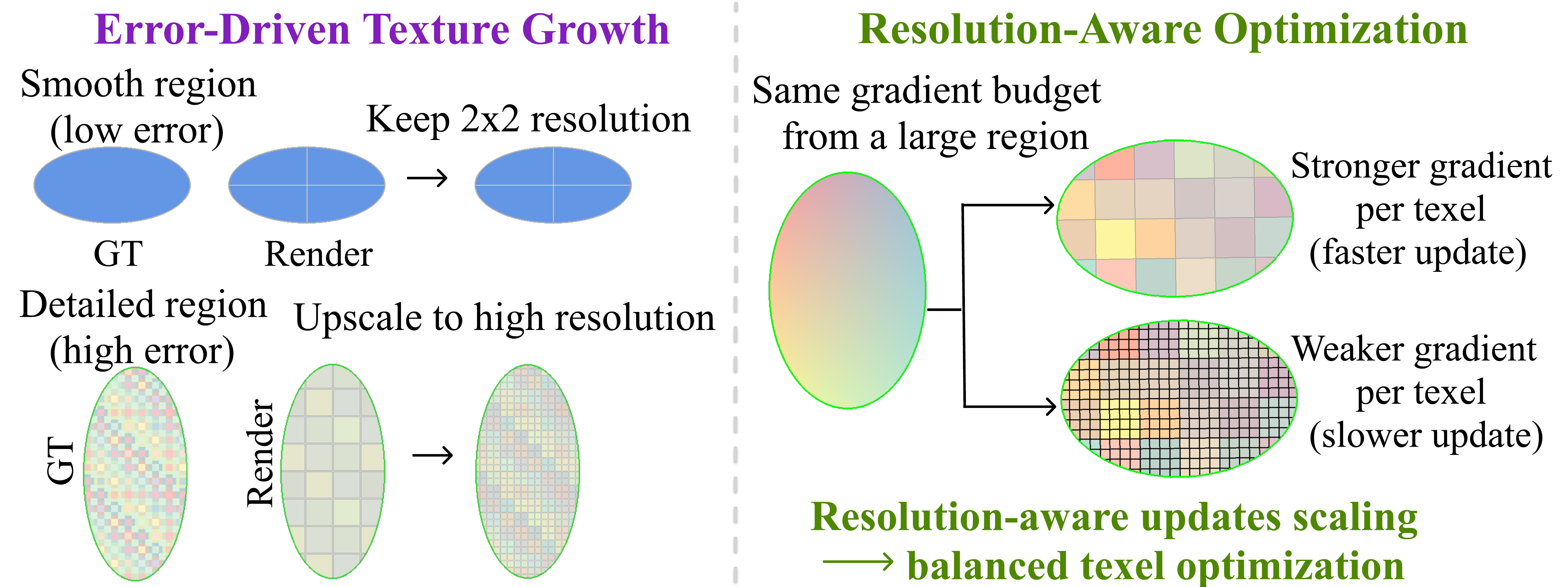}
    \caption{Left: Texture resolution is adaptively increased only for primitives with large reconstruction errors. Right: We use resolution-aware optimization to mitigate gradient dilution caused by texture upsampling.}
    \label{method}
\end{figure}

\subsection{Adaptive Per-Gaussian Texture Optimization}

Once the resolution schedule is in place, the next question is how to decide which primitives should receive additional texel capacity. We answer this by maintaining per-primitive statistics over the rendered training views. For each primitive, we track the projected area, the rendering contribution, and a normalized reconstruction error. The reconstruction error is computed from a weighted combination of $\ell_1$ and SSIM losses on the rendered image:
\begin{equation}
L_v(\mathbf{p}) =
(1-\lambda)
\left\| I_v(\mathbf{p}) - \hat{I}_v(\mathbf{p}) \right\|_1
+
\lambda
\bigl(1 - \mathrm{SSIM}_v(\mathbf{p})\bigr),
\end{equation}
where \(\mathrm{SSIM}_v(\mathbf{p})\) denotes the local SSIM map evaluated around pixel \(\mathbf{p}\), and \(\lambda\) balances the photometric and structural terms. We set \(\lambda=0.2\) in all experiments.
This image-space loss is assigned to primitives using the same blending weights produced by the rasterizer. For each primitive, we accumulate three statistics over the training views: projected area, rendering contribution, and reconstruction error. The projected area counts the number of pixels where the primitive has valid rasterization support. The rendering contribution measures its effective influence on the final image after alpha compositing: for a pixel $\mathbf{p}$ in view $v$, the contribution weight is $\omega_i^v(\mathbf{p})=T_i^v(\mathbf{p})\alpha_i^v(\mathbf{p})$, where $\alpha_i^v(\mathbf{p})$ is the opacity-weighted Gaussian response and $T_i^v(\mathbf{p})$ is the accumulated transmittance before primitive $i$ in the front-to-back order. Thus, a primitive can have low contribution because it is occluded by earlier primitives, has a weak opacity response, or affects only a small number of pixels. We then normalize the accumulated area and contribution by image size:
\begin{equation}
a_i =
\sum_v
\frac{n_{i,v}}{W_v H_v},
\qquad
c_i =
\sum_v
\frac{
\sum_{\mathbf{p} \in \Omega_v}
\omega_i^v(\mathbf{p})
}{
W_v H_v
}.
\end{equation}
where $n_{i,v}$ is the number of pixels covered by primitive $i$ in view $v$, and $W_vH_v$ is the image size. The accumulated error is normalized by contribution, giving a per-primitive normalized error
\begin{equation}
\tilde{e}_i = \frac{e_i}{c_i + \epsilon},
\end{equation}
where $e_i$ is the accumulated error and $\epsilon$ is a small constant for numerical stability. This normalization prevents primitives that are frequently visible but only weakly responsible for the final image from dominating the adaptation process.

The normalized error determines which primitives are allowed to increase their personal resolution ceiling. We define a high-error subset using a strict upper quantile:
\begin{equation}
\mathcal{H} =
\left\{
i \,\middle|\,
\tilde{e}_i > Q_q\left(\{\tilde{e}_j\}_{j=1}^{N}\right)
\right\},
\end{equation}
where $Q_q(\cdot)$ denotes the $q$-quantile of the normalized per-primitive errors.
In our implementation, we use $q=0.85$ for Mip-NeRF 360 and DeepBlending, selecting the top 15\% highest-error primitives for texture growth, and $q=0.60$ for Tanks\&Temples, selecting the top 40\%.
We found that a slightly larger growth ratio improves detail recovery on Tanks\&Temples, which contains more thin structures and localized high-frequency regions.
A sensitivity analysis of $q$ is provided in the supplementary material.

For every $i \in \mathcal{H}$, we increase the personal maximum texture resolution under the current scheduler-imposed texture cap:
\begin{equation}
R_{\max,i}
\leftarrow
\min\left(2R_{\max,i},\, u(t)\right),
\qquad i \in \mathcal{H}.
\end{equation}
Here, $u(t)$ is the scheduler-imposed global texture-resolution cap, while $R_{\max,i}$ is the personal resolution ceiling of primitive $i$.
This personal ceiling is a key part of the method. It makes texture growth primitive-specific instead of globally uniform: the scheduler-imposed cap $u(t)$ determines the maximum texture resolution allowed at the current stage, while the per-primitive reconstruction error determines which Gaussians can increase their own ceiling $R_{\max,i}$. This two-level design releases texture capacity progressively at the global level and consumes it selectively at the local level. As a result, difficult primitives can continue to expand when their reconstruction error remains high, whereas easy primitives remain compact and avoid unnecessary high-resolution allocation. This adaptive allocation behavior is illustrated in the left part of Figure~\ref{method}, where high-error primitives are selectively upsampled while smooth low-error regions remain compact.

In practice, texture-growth decisions are checked at every densification interval. We refresh the accumulated per-primitive statistics, increase the personal resolution ceilings of selected high-error primitives under the current global texture cap, and then recompute their active texture resolutions according to the current support, texel size, and personal ceiling. If the recomputed resolution exceeds the current texture size, the texture map is upsampled at the same structure-update step. The statistics are accumulated globally rather than maintained in a sliding window, while newly created primitives start with zero accumulated statistics. When a texture is upsampled, the learned texels are copied to initialize the enlarged texture grid; the Adam moments of existing texels are preserved, whereas newly introduced texels are initialized with zero moments. This preserves previously learned appearance while allowing newly allocated texture capacity to be optimized stably.

\subsection{Contribution- and Area-Aware Geometric Management}

Using the accumulated contribution $c_i$ and projected area $a_i$ defined above, we prune primitives that remain visually inactive over time. At each pruning check $t$, a primitive is marked as low-utility when
\begin{equation}
m_i^{(t)} =
\mathbb{1}\left[
c_i^{(t)} < \tau_c
\;\vee\;
a_i^{(t)} < \tau_a
\right],
\end{equation}
where $\tau_c$ and $\tau_a$ are the contribution and projected-area thresholds. We set $\tau_c=5\times10^{-6}$ and $\tau_a=5\times10^{-5}$ in all experiments. Since contribution and visibility may fluctuate across viewpoints and training stages, we maintain a patience counter for each primitive instead of pruning it immediately. The counter is increased whenever $m_i^{(t)}=1$ and reset otherwise; a primitive is removed only after being marked as low-utility for $K$ consecutive pruning checks, with $K=8$ by default. This temporal filtering avoids deleting primitives that are only temporarily occluded or weakly visible, while still eliminating persistently redundant geometry.

Pruning is checked periodically together with primitive management, using the same densification interval of 250 iterations in our default setting. To further reduce the risk of damaging thin structures, we do not rely on a single instantaneous contribution estimate. Instead, the patience mechanism requires a primitive to remain low-utility across consecutive checks before removal. In addition, newly split primitives are protected by a lower post-split opacity floor, which prevents them from being immediately removed before they have sufficient time to accumulate reliable contribution and area statistics. 

This contribution- and area-aware strategy is particularly suitable for texture-decoupled Gaussian representations. Once appearance details can be stored in local textures, not every primitive with nonzero opacity is necessary for reconstruction. By removing primitives that have consistently low effective contribution or negligible projected support, the model maintains a compact geometric scaffold and avoids spending texture parameters on visually inactive Gaussians.

\begin{table*}[t]
\centering
\scriptsize 
\footnotesize  
\setlength{\tabcolsep}{1.5pt} 
\renewcommand{\arraystretch}{1.0} 
\begin{tabular}{l|cccccc|cccccc|cccccc}
& \multicolumn{6}{c|}{Mip-NeRF 360} 
& \multicolumn{6}{c|}{Tanks\&Temples} 
& \multicolumn{6}{c}{DeepBlending} \\
& SSIM$\uparrow$ & PSNR$\uparrow$ & LPIPS$\downarrow$  & Params & FPS & Time
& SSIM$\uparrow$ & PSNR$\uparrow$ & LPIPS$\downarrow$  & Params & FPS & Time
& SSIM$\uparrow$ & PSNR$\uparrow$ & LPIPS$\downarrow$  & Params & FPS & Time \\
\hline 
CAT
&\cellcolor{orange!30}0.797  &\cellcolor{pink!70}27.15  &0.272    &\cellcolor{orange!30}135.0M  &\cellcolor{orange!30}46  &\cellcolor{yellow!40}90
&0.834  &23.45  &\cellcolor{yellow!40}0.232    &\cellcolor{orange!30}34.4M  &\cellcolor{orange!30}120  &\cellcolor{orange!30}34  
&\cellcolor{orange!30}0.909  &\cellcolor{pink!70}30.02  &\cellcolor{orange!30}0.308    &\cellcolor{orange!30}52.2M  &\cellcolor{orange!30}70  &\cellcolor{orange!30}53 \\
BBSplat
&0.783  &26.69  &\cellcolor{pink!70}0.233   &257.0M  &21  &115
&\cellcolor{orange!30}0.847  &\cellcolor{orange!30}23.58  &\cellcolor{pink!70}0.152    &324.3M  &\cellcolor{yellow!40}49  &50 
&0.895  &29.01  &\cellcolor{pink!70}0.260    &173.0M  &\cellcolor{yellow!40}25  &77 \\
GSTex
&\cellcolor{pink!70}0.802  &\cellcolor{yellow!40}27.01  &0.289   &\cellcolor{yellow!40}140.5M  &11  &\cellcolor{orange!30}66
&\cellcolor{pink!70}0.848  &\cellcolor{yellow!40}23.50  &0.239    &\cellcolor{yellow!40}78.9M &13  &\cellcolor{yellow!40}48 
&\cellcolor{pink!70}0.917  &\cellcolor{yellow!40}29.60  &\cellcolor{yellow!40}0.320    &\cellcolor{yellow!40}117.2M  &13  &\cellcolor{yellow!40}62\\
Textured-GS$_{10\%}$
&0.760  &26.63  &\cellcolor{yellow!40}0.260    &495.5M  &\cellcolor{yellow!40}42  &110
&0.805  &23.10  &0.240    &1494.0M  &21  &171 
&0.888  &28.79  &0.334    &1096.0M  &22  &153\\
Textured-GS$_{20\%}$
&0.762  &26.75  &\cellcolor{orange!30}0.256    &602.6M  &30  &122
&0.806  &23.06  &0.238    &1565.5M  &20  &174 
&0.891  &29.04  &0.323    &1400.1M  &24  &150\\
Ours
&\cellcolor{yellow!40}0.792  &\cellcolor{orange!30}27.13  &0.285    &\cellcolor{pink!70}61.2M  &\cellcolor{pink!70}62  &\cellcolor{pink!70}45 
&\cellcolor{yellow!40}0.845  &\cellcolor{pink!70}23.84  &\cellcolor{orange!30}0.211    &\cellcolor{pink!70}31.2M  &\cellcolor{pink!70}167  &\cellcolor{pink!70}20  
&\cellcolor{yellow!40}0.901  &\cellcolor{orange!30}29.74  &0.337    &\cellcolor{pink!70}19.8M  &\cellcolor{pink!70}110  &\cellcolor{pink!70}31 \\
\end{tabular}
\caption{
Quantitative comparison against state-of-the-art textured Gaussian methods under standard unconstrained training settings.
We report rendering quality metrics (SSIM, PSNR, LPIPS), efficiency metrics (Params, FPS), and training time in minutes.
The best results are highlighted in \colorbox{pink!70}{pink}, the second-best results in \colorbox{orange!30}{orange}, and the third-best results in \colorbox{yellow!40}{yellow}.
}
\label{table1} 
\end{table*}

\subsection{Resolution-Aware Optimization}

A subtle but important issue appears when texture maps are resized dynamically. As texture resolution increases, the number of texels grows quadratically, but the amount of image-space supervision does not grow at the same rate. This leads to gradient dilution: each texel receives a smaller effective update, and high-resolution textures tend to converge more slowly than low-resolution ones. To mitigate this issue, we introduce a resolution-aware update rule for the texture branch.

Let $\Delta \theta_i^{\mathrm{Adam}}$ be the standard Adam~\cite{kingma2014adam} update for the texture parameters of primitive $i$. We compute an effective update as
\begin{equation}
\Delta \theta_i^{\mathrm{eff}}
=
\lambda_i\,\Delta \theta_i^{\mathrm{Adam}},
\qquad
\lambda_i =
\operatorname{clip}
\left(
\left(\frac{A_i}{A_0}\right)^{\gamma},
1,
\lambda_{\max}
\right),
\end{equation}
where $A_i = H_i W_i$ is the current texture area of primitive $i$, $A_0=4$ corresponds to the initial $2\times2$ texture, $\gamma$ controls the compensation strength, and $\lambda_{\max}$ bounds the maximum update scaling. We set $\gamma=0.5$ and $\lambda_{\max}=4$ in all experiments.

This scaling is motivated by the fact that bilinear sampling distributes the backpropagated signal from each pixel to neighboring texels. After texture upsampling, the same image-space supervision is spread over more texels, which can reduce the effective per-texel update. The factor $\lambda_i$ empirically compensates for this dilution effect by increasing the effective update magnitude for higher-resolution textures.

The resolution-aware update is implemented in our custom fused Adam optimizer. Specifically, the per-element scale is applied only to the texture-map parameter group during the fused optimizer step, while the Gaussian geometry, opacity, scale, and rotation parameters are optimized with their original update rules. The Adam moment estimates are preserved, and the scale factor is used only to rebalance the effective texture update. Therefore, the optimizer remains aware of the current texture resolution without perturbing the optimization dynamics of the geometric scaffold. This mechanism makes adaptive texture growth practical: higher-resolution textures are not only allocated to difficult primitives, but also optimized at a sufficient rate after upsampling, as illustrated in the right part of Figure~\ref{method}.

\begin{table*}[t]
\centering
\scriptsize 
\footnotesize  
\setlength{\tabcolsep}{2.0pt} 
\renewcommand{\arraystretch}{1.1} 
\begin{tabular}{l|cccccc|cccccc|cccccc}
& \multicolumn{6}{c|}{Mip-NeRF 360} 
& \multicolumn{6}{c|}{Tanks\&Temples} 
& \multicolumn{6}{c}{DeepBlending} \\
& SSIM$\uparrow$ & PSNR$\uparrow$ & LPIPS$\downarrow$  & Params &FPS & Time
& SSIM$\uparrow$ & PSNR$\uparrow$ & LPIPS$\downarrow$  & Params &FPS & Time
& SSIM$\uparrow$ & PSNR$\uparrow$ & LPIPS$\downarrow$  & Params &FPS & Time \\
\hline 
CAT
&\cellcolor{orange!30}0.767  &\cellcolor{orange!30}26.51  &0.305    &61.2M  &48  &72
&0.820  &\cellcolor{orange!30}23.18  &\cellcolor{orange!30}0.256    &31.2M  &131  &30  
&0.896  &\cellcolor{orange!30}29.56  &0.351    &19.8M  &70  &50 \\
BBSplat
&0.739  &25.13  &\cellcolor{orange!30}0.297    &61.2M  &36  &54
&0.716  &21.07  &0.347    &31.2M  &117  &\cellcolor{pink!70}19 
&0.862  &27.18  &\cellcolor{orange!30}0.346    &19.8M  &67  &36 \\
GSTex
&0.732  &25.80  &0.395    &61.2M  &29  &\cellcolor{pink!70}40
&\cellcolor{orange!30}0.821  &22.94  &0.298    &31.2M  &33  &23 
&\cellcolor{pink!70}0.904  &28.74  &0.367    &19.8M  &55  &\cellcolor{orange!30}33\\
Textured-GS
&0.740  &26.18  &0.316    &61.2M  &\cellcolor{pink!70}117  &81
&0.783  &22.56  &0.270    &31.2M  &\cellcolor{orange!30}139  &59
&0.881  &28.76  &0.377    &19.8M  &\cellcolor{pink!70}124  &78\\
Ours
&\cellcolor{pink!70}0.792  &\cellcolor{pink!70}27.13  &\cellcolor{pink!70}0.285    &61.2M  &\cellcolor{orange!30}62  &\cellcolor{orange!30}45 
&\cellcolor{pink!70}0.845  &\cellcolor{pink!70}23.84  &\cellcolor{pink!70}0.211    &31.2M  &\cellcolor{pink!70}167  &\cellcolor{orange!30}20  
&\cellcolor{orange!30}0.901  &\cellcolor{pink!70}29.74  &\cellcolor{pink!70}0.337    &19.8M  &\cellcolor{orange!30}110  &\cellcolor{pink!70}31 \\
\end{tabular}
\caption{
Quantitative comparison under matched parameter budgets.Each baseline is configured with its native capacity-control mechanism to match the parameter count of our method as closely as possible.
The best results are highlighted in \colorbox{pink!70}{pink}, and the second-best results are highlighted in \colorbox{orange!30}{orange}.Under comparable model sizes, our method better preserves rendering quality, indicating more effective allocation of the available geometry and texture capacity.
}
\label{table2} 
\end{table*}

\begin{figure*}[t]
  \centering
  \includegraphics[width=1\linewidth]{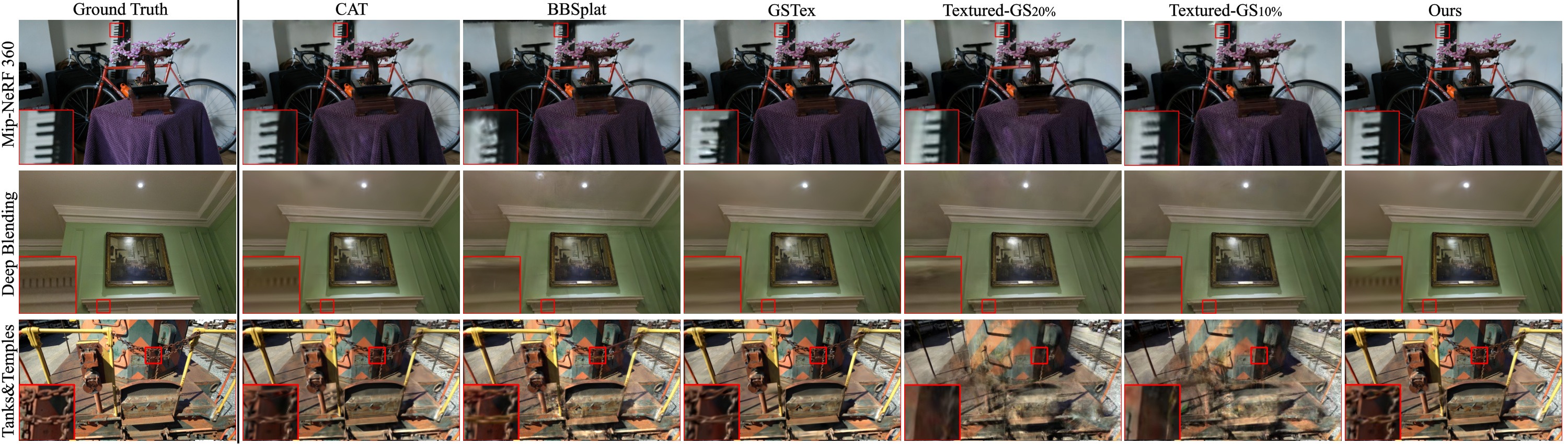}
  \caption{Our approach preserves high-frequency details while using fewer parameters and shorter training time than the compared textured Gaussian baselines.}
  \label{fig2}
\end{figure*}

\subsection{Training Procedure}

The full training loop alternates between ordinary optimization and periodic structure updates. At regular intervals, the scheduler returns the current render scale, global texture resolution cap, and densification rate. The current render scale is used to determine the resolution of the supervision images, while the global texture resolution cap determines the maximum allowed texture resolution at that stage. During the same interval, we refresh the per-primitive statistics from the training views and use them to decide which primitives should grow their personal ceiling, which primitives should be upsampled, which primitives should be split, and which primitives should be pruned. The render-scale statistics are also tracked so that the texel-to-pixel comparison remains consistent when supervision temporarily happens at a reduced resolution.

Overall, the method couples representation, scheduling, and optimization into a single content-aware training loop. Smooth regions remain compact, difficult regions receive more texture capacity, elongated primitives are refined geometrically when necessary, and unimportant primitives are removed. At the same time, the optimizer compensates for the changing texture resolution so that newly allocated capacity can be learned efficiently. This coordinated design is the central reason why the method can preserve both compactness and visual fidelity throughout training.

\begin{figure*}[t]
  \centering
  \includegraphics[width=1\linewidth]{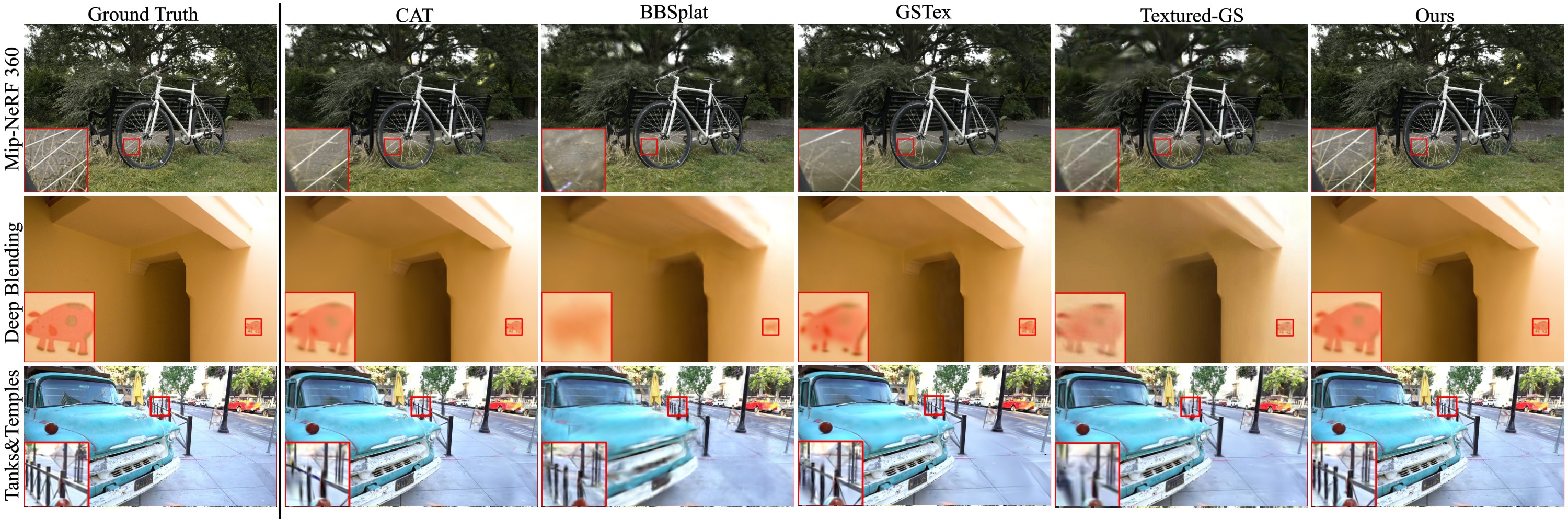}
  \caption{Qualitative comparisons under matched parameter budgets. When configured to use comparable parameter counts, fixed or heuristic allocation baselines tend to lose high-frequency details, whereas our adaptive allocation strategy better preserves local structures and texture details.}
  \label{fig3}
\end{figure*}

\section{Experiments}

\subsection{Experimental Setup}

\textbf{Datasets.} We evaluate our proposed method and state-of-the-art baselines on three widely adopted datasets for novel view synthesis: Mip-NeRF 360~\cite{barron2022mip}, Tanks and Temples~\cite{knapitsch2017tanks}, and Deep Blending~\cite{hedman2018deep}. Specifically, we report results on all nine scenes from the challenging Mip-NeRF 360 dataset. Unlike some prior works that use downsampled versions, we aggressively benchmark our model by training and evaluating on the original full-resolution images for Mip-NeRF 360. For the Tanks and Temples dataset, we select the large-scale \textit{train} and \textit{truck} scenes. For the Deep Blending dataset, we evaluate on the highly detailed \textit{playroom} and \textit{drjohnson} scenes. Following the standard evaluation protocol established by 3D Gaussian Splatting~\cite{kerbl20233d}, we construct our test set by holding out every 8th image in the sequence, utilizing the remaining images for optimization.

\noindent\textbf{Evaluation Metrics.}
We evaluate rendering quality using three standard metrics: Peak Signal-to-Noise Ratio (PSNR), Structural Similarity Index (SSIM), and Learned Perceptual Image Patch Similarity (LPIPS). We also report resource and efficiency metrics, including total parameter count, rendering speed in frames per second (FPS), and training time in minutes. It is worth noting that the original implementations of GSTex~\cite{rong2025gstex} and Textured-GS~\cite{chao2025textured} report LPIPS scores computed with AlexNet, which typically yields lower numerical values than the VGG-based LPIPS setting. Since the other baselines already report VGG-based LPIPS by default, we only re-evaluate the rendered test images of GSTex and Textured-GS using the VGG-based LPIPS metric~\cite{zhang2018unreasonable} for consistency.

\noindent\textbf{Implementation Details.}
Our method is implemented in PyTorch with a custom CUDA rasterization pipeline. Following standard practice, all scenes are optimized for 30,000 iterations. For the unconstrained comparison, CAT, BBSplat, and GSTex are reproduced using their official implementations and default training settings. For Textured-GS, we also follow its official implementation and default pipeline. Since the native 50$\times$50 texture configuration exceeds the available GPU memory on several large scenes, we use a memory-feasible texture resolution for these scenes while keeping all other optimization settings, including the training schedule, loss functions, optimizer configuration, and iteration budget, unchanged. The complete per-scene Textured-GS configuration is reported in the supplementary material. Under the unconstrained setting, Textured-GS optimizes RGBA texture channels following its original design. In the matched-budget comparison, all baselines are adjusted only through their native capacity-control mechanisms to approximately match the total representation size of our method; no post-hoc compression, additional pruning, loss modification, or method-specific hyperparameter tuning is introduced. All methods are evaluated using the same train/test splits, metrics, evaluation scripts, and hardware environment. All experiments are conducted on a single NVIDIA A800 GPU to ensure consistent training-time and FPS measurements.

\begin{figure*}[t]
  \centering
  \vspace{-2mm}
  \includegraphics[width=1\linewidth]{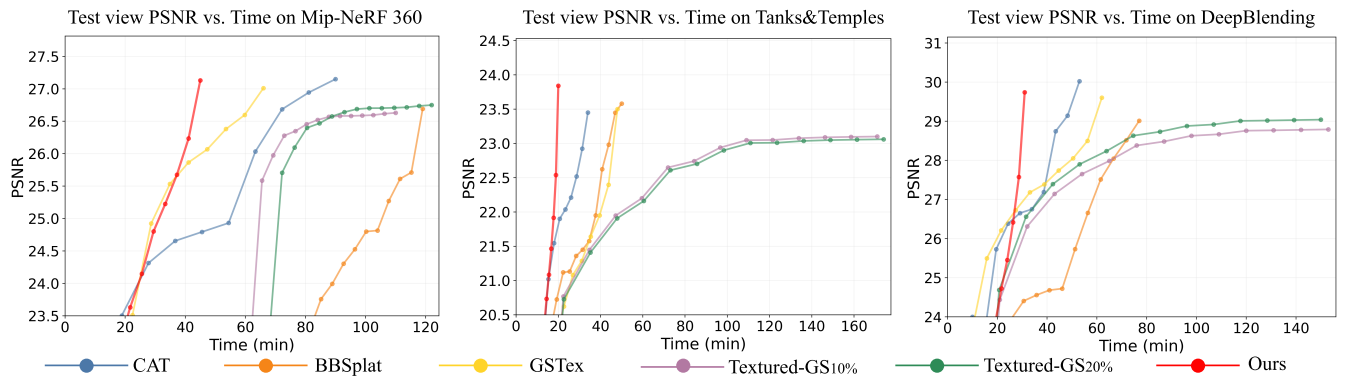}
  \vspace{-1.5mm}
  \caption{We present a convergence analysis of test-view PSNR versus training time. Our method achieves higher PSNR with shorter training time.}
  \label{fig4}
\end{figure*}

\subsection{Comparison with State-of-the-Art}

We compare our approach against representative textured Gaussian methods, including CAT~\cite{papantonakis2025content}, BBSplat~\cite{svitov2025billboard}, GSTex~\cite{rong2025gstex}, and Textured-GS~\cite{chao2025textured}. For Textured-GS, we report two variants because its texturing stage operates on a retained subset of primitives from the first-stage 2DGS reconstruction. Textured-GS$_{10\%}$ follows the default retention ratio and produces a primitive count comparable to ours, whereas Textured-GS$_{20\%}$ retains more primitives and produces a primitive count comparable to CAT, serving as a higher-capacity variant.
As reported in Table~\ref{table1} and visualized in Figure~\ref{fig3}, under standard unconstrained training settings, LiteTex-GS achieves a favorable quality--efficiency trade-off rather than optimizing for the best score on every metric. On Mip-NeRF 360, it obtains PSNR comparable to CAT while using substantially fewer parameters and shorter training time; on Tanks\&Temples, it achieves the best PSNR with the smallest parameter count and shortest training time; on DeepBlending, although CAT and GSTex achieve higher scores on some metrics, LiteTex-GS remains competitive while reducing representation size and training cost. These results indicate that our adaptive capacity allocation and geometric management improve the quality--efficiency trade-off by maintaining competitive visual quality while substantially reducing representation size and training cost. Beyond the final quality--efficiency trade-off, we also evaluate training convergence in Figure~\ref{fig4}, showing that LiteTex-GS reaches competitive PSNR within a shorter wall-clock time, whereas several baselines require longer optimization to approach their final performance. This suggests that the frequency-aware scheduler and resolution-aware optimization not only reduce the final model size, but also improve the quality--time trade-off during training.
\begin{table*}[t]
\centering
\vspace{-2mm}
\normalsize
\setlength{\tabcolsep}{6.0pt}
\begin{tabular}{lccccc cccccc}
\hline
ID & Scheduler & Pruning & Growth & Update & SSIM$\uparrow$ & PSNR$\uparrow$ & LPIPS$\downarrow$ & Points & Texels & Params & Time \\
\hline
I   &  &  &  &  & 0.797 & 27.15 & 0.272 & 255K & 40.0M & 135.0M & 90 \\
II  & \checkmark &  &  &  & 0.783 & 26.92 & 0.298 & 162K & 14.1M & 52.0M  & 50 \\
III & \checkmark & \checkmark &  &  & 0.779 & 26.80 & 0.300 & 119K & 16.2M & 55.8M  & 45 \\
IV  & \checkmark & \checkmark & \checkmark &  & 0.780 & 26.86 & 0.305 & 173K & 10.5M  & 41.7M  & 45 \\
V   & \checkmark & \checkmark & \checkmark & \checkmark & 0.792 & 27.13 & 0.285 & 209K & 16.3M & 61.2M & 45 \\
\hline
\end{tabular}

\vspace{-2mm}

\caption{
Component ablation on the Mip-NeRF 360 dataset. The scheduler and pruning reduce redundant capacity and training time, error-driven texture growth reallocates texture capacity toward high-error regions, and resolution-aware update scaling improves the optimization of upsampled textures without increasing training time.
}
\label{tab:ablation_components}
\end{table*}

\subsection{Comparisons under Matched Parameter Budgets}

Default configurations of textured Gaussian methods often lead to substantially different parameter counts, since each method distributes geometric and textural capacity in a different way. Therefore, unconstrained comparisons reflect the overall quality--efficiency trade-off of each complete pipeline, but they do not fully isolate how effectively each representation uses a comparable capacity budget. To further evaluate intrinsic representational efficiency, we conduct a matched-budget comparison in Table~\ref{table2} and Figure~\ref{fig3}, where each baseline is configured to use approximately the same representation capacity as our method. Under comparable parameter budgets, our method preserves reconstruction quality more effectively across the evaluated datasets. Fixed-resolution or heuristic allocation strategies tend to lose high-frequency details when the parameter budget is limited, whereas our method allocates texture capacity according to reconstruction error and removes low-utility geometry. These results suggest that the proposed content-adaptive allocation strategy uses the available geometric and textural capacity more effectively than fixed or heuristic allocation schemes.

For this setting, we adjust only capacity-related controls of each baseline while keeping its original training pipeline, loss functions, optimizer settings, and iteration budget unchanged. For CAT, GSTex, and Textured-GS, whose parameterization is directly comparable to ours, we align both the primitive count and the total number of RGB texels whenever possible. For Textured-GS, we use RGB textures in the matched-budget setting to make its texture-parameter definition consistent with CAT, GSTex, and our method, while its unconstrained setting follows the original RGBA design. For BBSplat, which uses a different RGBA texture parameterization, we derive the closest feasible configuration according to its native parameter formula. No additional pruning, post-hoc compression, loss modification, or method-specific hyperparameter tuning is introduced for any baseline. The detailed matching protocol and per-method capacity controls are provided in the supplementary material.

\subsection{Ablation Studies}

To validate our architectural decisions, we conduct component-wise ablations on the Mip-NeRF 360 dataset.
We ablate the main components of LiteTex-GS in Table~\ref{tab:ablation_components}. The baseline achieves reasonable rendering quality but relies on a large parameter footprint of 135.0M. Adding the frequency-aware scheduler reduces the total parameters from 135.0M to 52.0M and shortens the training time from 90 minutes to 50 minutes, showing that coarse-to-fine supervision suppresses redundant early-stage capacity. Contribution- and area-aware pruning further reduces the point count from 162K to 119K while keeping the same training time. With error-driven texture growth, the model reallocates texture capacity toward high-error regions, reducing the texel count from 16.2M to 10.5M and the total parameters from 55.8M to 41.7M, with PSNR improving from 26.80 to 26.86. Finally, adding resolution-aware update scaling improves PSNR from 26.86 to 27.13 and LPIPS from 0.305 to 0.285 without increasing training time. Although the final active texel count increases, this reflects a different optimized texture distribution induced by the update rule, rather than parameters directly introduced by the rule itself. We provide the corresponding texture-resolution distributions before and after applying this module in the supplementary material. Overall, these results show that the proposed scheduler, pruning, texture growth, and resolution-aware optimization jointly improve the quality--efficiency trade-off.

To isolate adaptive texture allocation from the inherited scheduling
principle, we combine DashGaussian-style scheduling with fixed
$4{\times}4$, $8{\times}8$, and $16{\times}16$ per-Gaussian textures.
Table~\ref{dash_fixed} shows that increasing the fixed texture
resolution improves reconstruction quality but rapidly increases the
parameter count, while our adaptive allocation achieves a more favorable
quality--efficiency trade-off.

\begin{table}[t]
\centering
\vspace{-3mm}
\normalsize
\setlength{\tabcolsep}{1.6pt}
\renewcommand{\arraystretch}{1.08}
\begin{tabular}{@{}llccccc@{}}
\toprule
Dataset & Texture
& PSNR$\uparrow$
& SSIM$\uparrow$
& LPIPS$\downarrow$
& Params$\downarrow$
& Time$\downarrow$ \\
\midrule

\multirow{4}{*}{Mip-NeRF 360}
& $4{\times}4$   & 25.14 & 0.696 & 0.420 & 4.7M  & 24 \\
& $8{\times}8$   & 25.38 & 0.715 & 0.381 & 10.1M & 24 \\
& $16{\times}16$ & 25.56 & 0.734 & 0.346 & 31.2M & 25 \\
& \textbf{Ours}           &27.13   &0.792 &0.285  &61.2M  &45    \\
\midrule

\multirow{4}{*}{DeepBlending}
& $4{\times}4$   & 28.03 & 0.871 & 0.405 & 3.1M  & 16 \\
& $8{\times}8$   & 28.28 & 0.876 & 0.391 & 6.8M  & 16 \\
& $16{\times}16$ & 28.23 & 0.881 & 0.370 & 21.7M & 17 \\
& \textbf{Ours}           &29.74  & 0.901 & 0.337 & 19.8M & 31   \\
\midrule
\multirow{4}{*}{Tanks\&Temples}
& $4{\times}4$   & 22.08 & 0.774 & 0.324 & 4.6M  & 11 \\
& $8{\times}8$   & 22.24 & 0.789 & 0.289 & 9.9M  & 12 \\
& $16{\times}16$ & 22.37 & 0.803 & 0.257 & 31.5M & 14 \\
& \textbf{Ours}           & 23.84 & 0.845 & 0.211 &  31.2M     &20    \\
\bottomrule
\end{tabular}
\vspace{-3mm}

\caption{
Comparison with DashGaussian-style scheduling using fixed
per-Gaussian textures.
}
\label{dash_fixed}
\vspace{-3mm}
\end{table}
\label{others}

\section{Conclusion}

In this work, we presented LiteTex-GS, a content-adaptive texturing framework for 2D Gaussian Splatting. By combining frequency-aware scheduling, contribution- and area-aware pruning, error-driven texture growth, and resolution-aware update scaling, our method adaptively allocates geometric and textural capacity during training. Experiments on standard novel view synthesis benchmarks show that LiteTex-GS achieves competitive rendering quality with substantially fewer parameters and shorter training time than existing textured Gaussian baselines.

\noindent\textbf{Limitations.}
Our method still has several limitations. The frequency-based scheduler
may be affected by spurious high-frequency signals caused by image noise
or repetitive texture patterns, which can bias the estimated frequency
distribution and reduce computational efficiency. The pruning strategy
may require careful threshold selection for thin or weakly visible
structures. Extending the framework to sparse-view inputs, dynamic scenes,
and challenging materials such as strong reflections or transparency
remains future work.

\vspace{-3mm}
\vspace{-2mm}
\section{Acknowledgments}
This work was supported in part by the Beijing Natural Science Foundation under Grant JQ24023 and Grant F251020, in part by Beijing Municipal Science \& Technology Commission Project under Grant Z231100006623010, and in part by Peking University Medicine plus X Pilot Program — Artificial Intelligence and Medical Development Initiative under Grant BMU2025YXXLHAIYX020.

\bibliographystyle{eg-alpha-doi} 
\bibliography{egbibsample}       


\end{document}